\documentclass[runningheads]{llncs}

\usepackage{eccvabbrv}

\usepackage{graphicx}
\usepackage{booktabs}
\usepackage{xcolor}        
\usepackage{graphicx}
\usepackage{booktabs}
\usepackage{multirow} 
\usepackage{subcaption}
\usepackage{pifont}
\usepackage[accsupp]{axessibility}  

\definecolor{myblue}{HTML}{2C7DD9}

\usepackage{hyperref}
\usepackage{ulem}
\usepackage{orcidlink}

\begin{document}

\title{MMDG-Bench: A Benchmark for Multimodal Domain Generalization} 

\titlerunning{MMDG-Bench}

\author{Qianshan Zhan\inst{1}  \and
Qian Wang\inst{2} \and
Da Li\inst{3} \and
Xiao-Jun Zeng\inst{1} \and
Xiatian Zhu \inst{4}}
\authorrunning{Q.~Zhan et al.}

\institute{University of Manchester \and
Jiyue AI  
\\ 
\and
Samsung AI Centre Cambridge\\
\and
University of Surrey}
\maketitle

\begin{abstract}

Multi-modal Domain Generalization (MMDG) seeks to leverage complementary modalities to enhance model robustness on unseen domains. Despite extensive progress in Multi-modal Learning (MML) and Domain Generalization (DG) as individual fields, their systematic integration remains under-explored. Current MMDG research is largely confined to action recognition and lacks standardized evaluation protocols. To address this, we introduce MMDG-Bench, a comprehensive benchmark featuring two foundational frameworks: ``DG then MML'' (D2M) and ``MML then DG'' (M2D). We provide unified experimental protocols across diverse tasks, including video-audio-flow action recognition and RGB-Depth-IR face anti-spoofing. By instantiating ten MMDG baselines through pairing a unified MML configuration with five DG techniques under both D2M and M2D orderings, we demonstrate that these structured combinations frequently outperform existing state-of-the-art methods, underscoring the necessity of a unified benchmarking effort. Our analysis yields three key insights: (1) Integrating DG techniques provides consistent generalization gains across various backbones, whereas non-DG methods are highly sensitive to backbone shifts; (2) The optimal framework choice depends on inter-modal stability: D2M excels when modal relations are stable across domains, while M2D is more robust to cross-domain relational variance; (3) Stronger backbones yield amplified performance dividends when integrated into our structured frameworks. MMDG-Bench provides a principled foundation and actionable design guidelines for future research in multi-modal robustness.
Code is released at \url{https://github.com/qszhan/MMDG-Bench}.

\end{abstract}

\section{Introduction}
\label{sec:intro}

Multi-modal domain generalization (MMDG)~\cite{dong2023simmmdg} leverages complementary information across modalities for robust generalization to unseen domains. MMDG addresses two interacting gaps: (1) the modality gap from heterogeneity across modalities, and (2) the domain gap from environmental and sensor variations. This interaction makes MMDG fundamentally more complex than single-modal domain generalization (DG)~\cite{gulrajani2020search} or intra-domain multi-modal learning (MML)~\cite{xu2023multimodal}.

While MML and DG have been extensively studied, existing MMDG research lacks systematic investigation of leveraging and evaluating these methods and their combination. Early MMDG studies focused on modality alignment and fusion~\cite{planamente2022domain,dong2023simmmdg,fan2024cross,wang2025modality,ji2026alignment} while overlooking the extensive MML literature and typically merging source domains at training. More recent efforts incorporated domain concepts but overlooked DG literature~\cite{huang2025bridging,lin2024suppress,li2025towards}. Furthermore, prior MMDG work focused exclusively on action recognition and lacked standardized evaluation protocols across tasks.

To address these limitations, we present MMDG-Bench, comprising two complementary frameworks: ``DG then MML'' (D2M) as illustrated in Fig.~\ref{fig_d2m_framework_sub} and ``MML then DG'' (M2D) as in Fig.~\ref{fig_m2d_framework_sub}, enabling systematic integration and evaluation of existing DG and MML methods. D2M learns domain-invariant representations per modality before cross-modal fusion; M2D aligns modalities then enhances domain generalization on fused features. We establish unified protocols for video-audio-flow action recognition and RGB-Depth-IR face anti-spoofing, implementing ten MMDG variants by pairing a unified MML configuration with five DG techniques. Our design is flexible and can accommodate additional methods and tasks.

Extensive experiments validate our frameworks, with variants that often surpass prior state-of-the-art results. Our analysis reveals key design principles for framework selection, DG integration, and practical deployment. This work establishes a standardized foundation for MMDG research through unified frameworks, rigorous benchmarking, and actionable guidance.


\section{Related Work}
\label{sec:2_related}

\textbf{Multi-modal Learning} leverages complementary information across modalities to learn richer and more robust representations than single-modality models. A major line of work focuses on feature alignment, where contrastive objectives pull paired representations together while preserving modality-specific cues, as shown in CLIP \cite{radford2021learning}, ALIGN \cite{jia2021scaling}, and translation-based models that reconstruct one modality from another to enforce cross-modal consistency \cite{dong2023simmmdg}. Beyond representation learning, optimization-based balancing methods such as Gradient Blending \cite{wang2020makes} address modality dominance by adaptively reweighting gradients. Fusion-oriented approaches study how to combine modalities at different stages, from early fusion \cite{owens2018audio} to late fusion \cite{yao2022modality}, with attention bottlenecks \cite{nagrani2021attention} emerging as a compact and effective solution for cross-modal interaction.
 

\noindent \textbf{Domain Generalization} aims to train models on multiple source domains that generalize to unseen targets \cite{li2017deeper}. DG approaches fall into three categories: data manipulation, representation learning, and learning strategy–based methods \cite{wang2022generalizing}.
Data manipulation enhance generalization by increasing diversity through techniques like Mixup \cite{zhang2017mixup}, which interpolates samples and labels to encourage smoother decision boundaries. Representation learning seeks domain-invariant features by minimizing statistical or semantic discrepancies, as in MMD \cite{gretton2006kernel} or cross-domain contrastive learning \cite{motiian2017unified}. Learning strategy–based methods improve generalization through meta-learning \cite{li2018learning} and self-supervision \cite{carlucci2019domain}.
Notably, MIRO \cite{cha2022domain} introduces a mutual-information regularization objective that enforces semantic consistency across domains, boosting generalization.

\noindent \textbf{Multi-modal Domain Generalization} aims to learn from multi-modal source domains and generalize to unseen targets. Early studies primarily focused on MML.  For action recognition, RNA-Net \cite{planamente2022domain} aligned audio–visual norms to enhance cross-modal robustness, while SimMMDG \cite{dong2023simmmdg} jointly learned shared and modality-specific features to improve alignment. CMRF \cite{fan2024cross} further flattened cross-modal manifolds to reduce modality bias. More recently, MBCD~\cite{wang2025modality} mitigates modality imbalance induced by naïve weight averaging via collaborative distillation, and MAD-DG~\cite{ji2026alignment} tackles asynchronous modality misalignment through segment-label aligned temporal binding.
These methods merge data from all source domains, implicitly assuming domain homogeneity and thus limiting generalization. 
More recent efforts explicitly incorporate DG into multi-modal learning. For action recognition, UR-Mixup \cite{huang2025bridging} augments mixed-domain data. Li et al. \cite{li2025towards} adversarially suppress domain-dependent patterns in both modality-specific and fused representations. For face anti-spoofing, Lin et al. \cite{lin2024suppress} introduce domain-specific prototypes.
Despite progress, existing approaches lack systematic frameworks integrating MML and DG, and most focus solely on action recognition without standardized protocols for fair comparison.

\section{MMDG-Bench}
\label{sec:3_bench}

\subsection{Problem Formulation}
Let $\mathcal{X} = \prod_{m=1}^M \mathcal{X}_m$ and $\mathcal{Y}$ denote the $M$-modal input space and the output space, respectively. 
We are given a set of $K$ source domains $\mathcal{S} = \{D_{s}^{(k)}\}_{k=1}^K$, where $D_{s}^{(k)} = \{(\mathbf{x}_i^{(k)}, y_i^{(k)})\}_{i=1}^{n_k}$ contains $n_k$ labeled instances.
Each instance $\mathbf{x} = \{ \mathbf{x}_m \}_{m=1}^M$ consists of $M$ modalities. 
The domains are subject to distribution shifts, i.e., $P_{XY}^{(i)} \neq P_{XY}^{(j)}$ for any $i \neq j$. 
The goal of MMDG is to learn a predictive function $f: \mathcal{X} \rightarrow \mathcal{Y}$ that achieves minimum risk on an unseen target domain $D_t$: $f = \arg \min_{f} \mathbb{E}_{(\mathbf{x},y) \sim D_t} \left[ \ell(f(\mathbf{x}), y) \right]$, 
where $\ell(\cdot, \cdot)$ denotes the loss function.
As $D_t$ is inaccessible during training, MMDG seeks domain and modality invariance via a surrogate objective over $\mathcal{S}$. Let $\{ \mathbf{z}^{(k)} \}_{k=1}^K$ denote features grouped by domain and $\{ \mathbf{z}_m \}_{m=1}^M$ denote features grouped by modality. The objective is defined as \(\min_{f} \sum_{k=1}^{K}  \mathbb{E}_{(\mathbf{x},y) \sim D_s^{(k)}} \left[ \ell(f(\mathbf{x}), y) \right] + \lambda_{d} \Omega_{DG}\left( \{ \mathbf{z}^{(k)} \}_{k=1}^K \right) + \lambda_{m} \Omega_{MML}\left( \{ \mathbf{z}_m \}_{m=1}^M \right)\), where the regularizer $\Omega_{DG}$ and $\Omega_{MML}$ enforce domain- and modality-invariance, and $\lambda_d, \lambda_m > 0$ weight their contributions.



%

\subsection{D2M and M2D Frameworks}

\begin{figure*}[!t]
\centering
\begin{subfigure}[t]{0.98\textwidth}
  \centering
  \includegraphics[width=\textwidth]{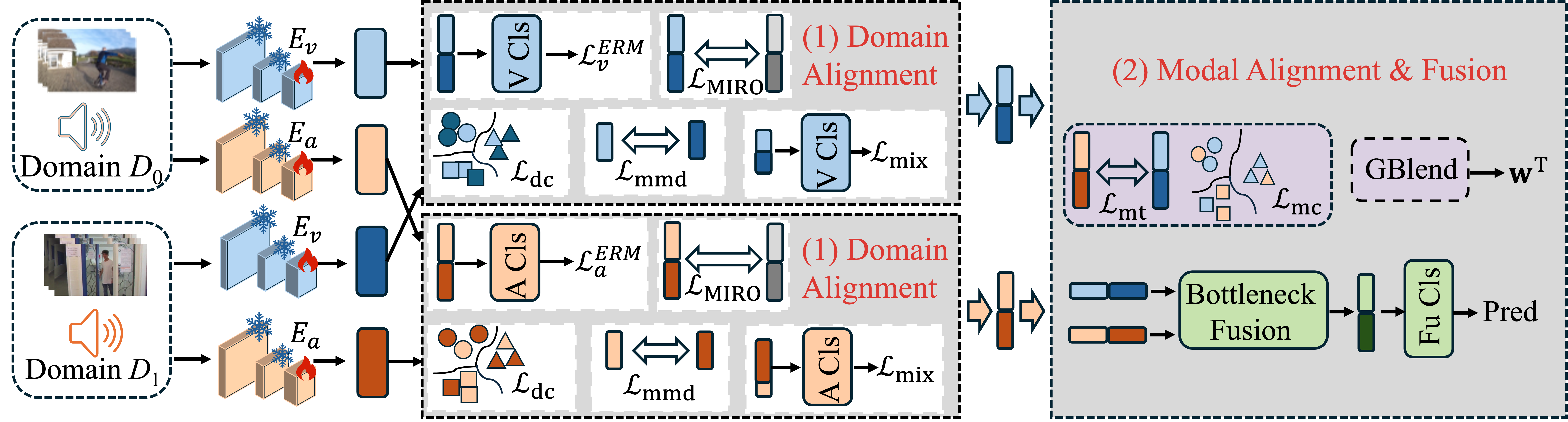}
  \caption{D2M: domain alignment first, modal fusion second.}
  \label{fig_d2m_framework_sub}
\end{subfigure}\hfill
\begin{subfigure}[t]{0.98\textwidth}
  \centering
  \includegraphics[width=\textwidth]{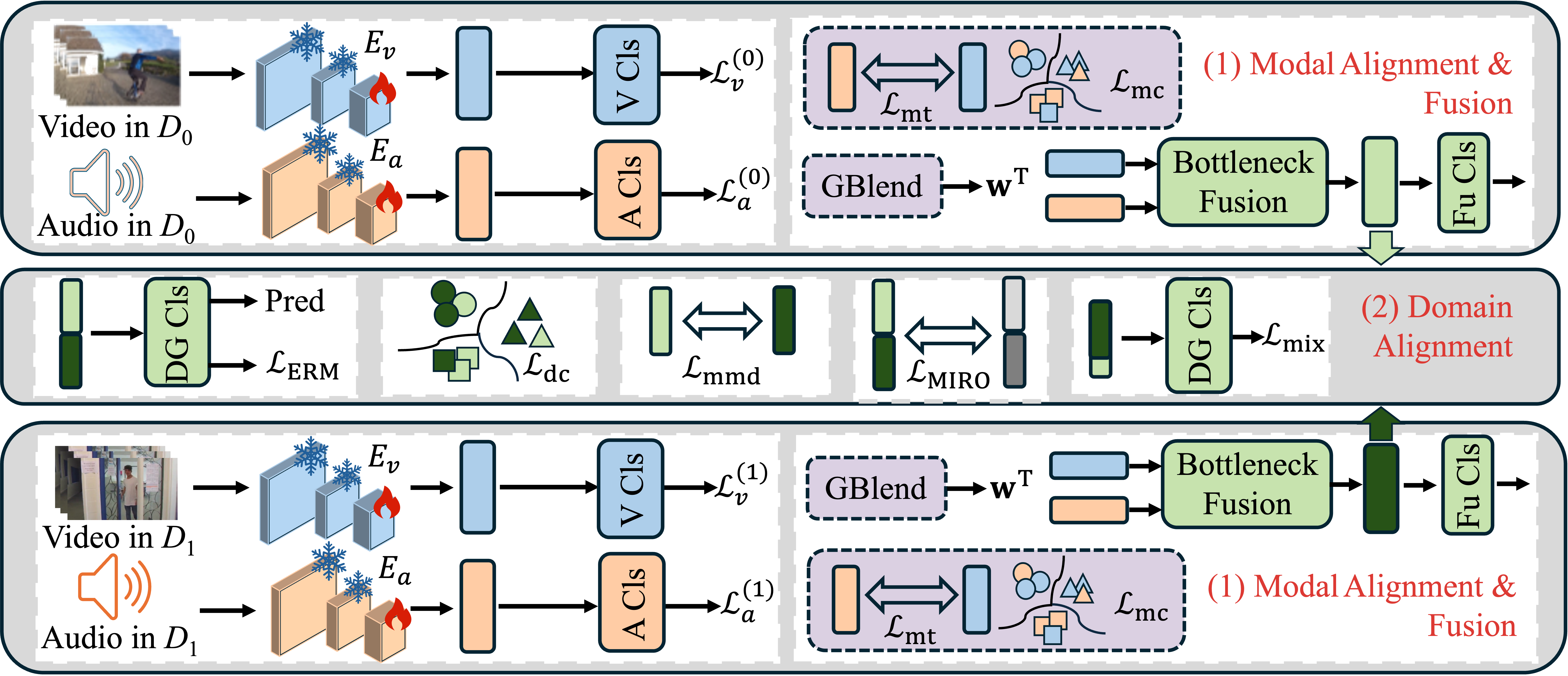}
  \caption{M2D: modal fusion first, domain alignment second.}
  \label{fig_m2d_framework_sub}
\end{subfigure}
\caption{\textbf{Two MMDG frameworks.}
Modal-specific encoders $E_v$ and $E_a$ extract features from domains $D_0$ and $D_1$, with partially frozen backbones and trainable upper layers.
MML aligns modalities via modality translation ($\mathcal{L}_{\text{mt}}$) and modality contrastive loss ($\mathcal{L}_{\text{mc}}$) before bottleneck fusion, while GBlend outputs modality-wise loss fusion weights $\mathbf{w}^T$.
DG uses ERM ($\mathcal{L}_{\text{ERM}}$) with optional regularizers: Mixup ($\mathcal{L}_{\text{mix}}$), MMD ($\mathcal{L}_{\text{mmd}}$), contrastive regularization ($\mathcal{L}_{\text{dc}}$), and MIRO ($\mathcal{L}_{\text{MIRO}}$).
\textbf{(a)} D2M first applies DG per modality across domains, then performs MML fusion on domain-invariant features.
\textbf{(b)} M2D first performs MML fusion within each domain, then applies DG on fused representations for final prediction.
}
\label{fig_frameworks}
\end{figure*}
We propose two complementary frameworks to combine MML and DG methods: DG then MML (D2M) and MML then DG (M2D), shown in Figs~\ref{fig_d2m_framework_sub} and \ref{fig_m2d_framework_sub}.
D2M first mitigates the domain gap by aligning each modality across domains through dedicated DG modules, then performs multi-modal fusion on the domain-invariant yet modality-specific representations. This ensures modality fusion operates on domain-invariant features, enhancing generalization to unseen domains.
M2D first mitigates the modality gap through multi-modal fusion within each source domain to capture complementary information, then aligns fused representations across domains to enhance domain invariance and improve generalization.

\subsection{MMDG Variants}

We consider the following representative MML and DG algorithms.

\textbf{MML algorithms.} We select four representative methods that cover diverse fusion strategies to comprehensively evaluate multi-modal learning approaches. \ding{172} Supervised Contrastive Learning \cite{dong2023simmmdg,fan2024cross} represents metric learning approaches that align cross-modal features under shared labels while preserving modality-specific cues. \ding{173} Modality Translation \cite{dong2023simmmdg} represents generative approaches that reconstruct one modality from another to enhance cross-modal consistency and mutual understanding. 
\ding{174} Gradient Blending \cite{wang2020makes} is an optimization-based method that adaptively reweights modality-specific gradients according to their overfitting–generalization behavior, thereby preventing modality dominance.
\ding{175} Attention Bottleneck Fusion \cite{nagrani2021attention} represents attention-based fusion that integrates modalities through a compact latent space capturing only the most informative cross-modal interactions. Together, these methods span metric learning, generative modeling, optimization balancing, and attention mechanisms—the primary paradigms in multi-modal learning.

\textbf{DG algorithms.} We select five methods spanning complementary generalization principles to systematically evaluate domain adaptation strategies. \ding{172} ERM \cite{vapnik1999overview} serves as the baseline with standard empirical risk minimization. \ding{173} Mixup \cite{zhang2017mixup} represents data augmentation approaches through linear interpolation between samples and domains. \ding{174} MMD \cite{gretton2006kernel} represents distribution matching methods that minimize statistical divergence between source domains. \ding{175} CL \cite{motiian2017unified} represents representation learning approaches that enforce semantic alignment through contrastive objectives. \ding{176} MIRO \cite{cha2022domain} represents information-theoretic methods using mutual information regularization for domain-invariant features. This selection covers the major DG paradigms: standard training, data augmentation, distribution alignment, representation learning, and information theory. All methods use SAM optimizer \cite{foret2020sharpness} to enhance flatness of loss landscapes for improved robustness.

\textbf{MMDG variants.} We perform exhaustive pairing between MML and DG methods under both D2M and M2D frameworks. 
ERM serves as the base training objective in all variants, representing standard empirical risk minimization for supervised learning, while additional DG methods (Mixup, MMD, CL, MIRO) augment ERM with domain-specific regularization. We use MM to denote the setting where all four MML components are enabled simultaneously. Fixing this unified MML setting keeps the variant space tractable and isolates the effect of the DG method. This yields five MML-DG configurations: \ding{172} MM+ ERM, \ding{173} MM+Mixup, \ding{174} MM+MMD, \ding{175} MM+CL, and \ding{176} MM+MIRO. Evaluating each under both frameworks produces ten variants, enabling the analysis of the interaction between DG regularization and framework ordering.
Detailed formulations and pseudo-code are in the Supplementary Material.

\section{Experiments}
With the constructed MMDG variants, we evaluate their performance under D2M and M2D frameworks on action recognition and face anti-spoofing tasks.

\subsection{MMDG Action Recognition} \label{sec_sc}
\textbf{Datasets.}
We evaluate on two multi-modal action recognition datasets: EPIC-Kitchens~\cite{munro2020multi} and Human-Animal-Cartoon (HAC)~\cite{dong2023simmmdg}, both providing video, optical flow, and audio modalities.
EPIC-Kitchens contains eight kitchen actions (put, take, open, close, wash, cut, mix, pour) recorded in three distinct kitchens, serving as domains (D1–D3).
HAC contains seven actions (sleeping, watching TV, eating, drinking, swimming, running, opening door) performed by humans, animals, and cartoons, defining domains (H, A, C).     \\
\textbf{Competitors.} We compare the ten MMDG variants under both frameworks against DeepAll~\cite{dong2023simmmdg}, which jointly trains on all source domains without an explicit DG strategy, and two action-recognition MMDG baselines, SimMMDG~\cite{dong2023simmmdg} CMRF~\cite{fan2024cross}, MDJAT \cite{li2025towards} and MAD-DG \cite{ji2026alignment}. For fair comparison, we re-implement these competitors in a unified setup, using the same backbone architecture and optimizer as our methods. We do not include RNA-Net~\cite{planamente2022domain} or UR-Mixup~\cite{huang2025bridging} because official code is not publicly available. \\
\textbf{Implementation.} 
Experiments are implemented with MMAction2~\cite{2020mmaction2}, following the settings described in~\cite{dong2023simmmdg}. We evaluate two backbone configurations for action recognition: a CNN-based setting and a transformer-based setting.
In the CNN-based setting, the video encoder is SlowFast~\cite{feichtenhofer2019slowfast} pretrained on Kinetics-400~\cite{kay2017kinetics}. The audio branch uses a ResNet-18~\cite{he2016deep} initialized from VGGSound~\cite{chen2020vggsound}. The optical-flow encoder adopts the slow-only pathway of SlowFast, also initialized from Kinetics-400. The resulting uni-modal feature dimensions are 2304 (video), 512 (audio), and 2048 (flow). Each is projected via a two-layer MLP (2048 hidden units, 128-dimensional output) for modal translation and modal contrastive losses.
In the transformer-based setting, we replace these CNN backbones with VideoMAEv2~\cite{wang2023videomaev2} (ViT-Base) for video and optical flow, and Audio Spectrogram Transformer~\cite{gong2021ast} (AST, ViT-Base) for audio. In this case, all modalities produce 768-dimensional features, removing the need for the projection layers used in the CNN setting.
For both settings, the domain classifier built on fused features is a two-layer MLP with 256 hidden units, while the modality fusion classifier follows the same structure with a hidden size equal to half of the input dimension. Optimization uses Sharpness-Aware Minimization (SAM)~\cite{foret2020sharpness} with AdamW~\cite{loshchilov2017decoupled} as the base optimizer and perturbation radius $\rho{=}0.05$. 
All experiments are conducted on a single NVIDIA H100 (80 GB) GPU, with an average runtime of approximately 1 hour per task for CNN-based setting and 5 hours for ViT-based setting. Further implementation details, as well as results for the two-modality cases, are provided in the Supplementary Material.

\begin{table*}[!htbp]
\centering
\setlength{\tabcolsep}{1.5pt}
\small
\caption{MMDG results with two backbone settings (CNN and ViT-Base) using video, audio and optical flow for action recognition. \uline{Underline}: The best result among the re-implemented models for a fair comparison.
\textbf{Bold}: The results surpassing the best-reproduced result in bold.
Results for two-modality cases (video+audio, video+flow, and audio+flow) are provided in the Supplementary Material.
}
\label{tab_hac_epic_all}
\begin{tabular}{@{}l l c c c c c c c c @{}}
\toprule[1pt]
& \multirow{2}{*}{Method} &
\multicolumn{4}{c}{HAC Dataset} &
\multicolumn{4}{c}{EPIC-Kitchens Dataset} \\
\cmidrule(lr){3-6}\cmidrule(lr){7-10} 
 &  
 & A,C$\to$H & H,A$\to$C & H,C$\to$A & Mean
 & 2,3$\to$1 & 1,3$\to$2 & 1,2$\to$3 & Mean \\\midrule[1pt]
\multicolumn{10}{c}{CNN-Based Backbone} \\ \midrule[1pt]
\multirow{4}{*}{} 
&DeepAll   \cite{dong2023simmmdg}    & 72.96 & 49.63 & 73.95 & 65.52  & 55.7 & 66.00 & 62.42 & 62.04 \\
&SimMMDG    \cite{dong2023simmmdg}   & 74.62 & 52.02 & 72.63 & 66.42 & \uline{63.21} & 67.73 & 63.45 & \uline{64.80} \\
&CMRF  \cite{fan2024cross}  & 74.69 & 47.79 & \uline{77.04} & 66.51 & 55.4 & 69.2 & \uline{67.76} & 64.12  \\
&MDJAT \cite{li2025towards}  & \uline{76.28} & \uline{54.10} & 69.97 & \uline{66.78} & 62.99 & \uline{70.80} & 60.47 & 64.75 \\
&MAD-DG \cite{ji2026alignment} 
& 74.27 & 53.17 & 71.02 & 66.15 
& 60.70 & 65.06 & 62.42 & 62.73 \\
\midrule
\multirow{5}{*}{D2M} 
&MM+ERM    &  \textbf{78.3}	&  52.48	&  \textbf{79.57}	& \textbf{70.17}  &61.84 &   70.53 & 66.63 &   \textbf{66.33} \\
&MM+Mixup     &  \textbf{81.54}	&47.89	&   \textbf{77.92}	&  \textbf{69.12} &   \textbf{63.45} &   \textbf{71.47} & 66.02 &   \textbf{66.98} \\
&MM+MMD     &   \textbf{78.44}	&    \textbf{56.53}	&  \textbf{79.47}	&  \textbf{71.48} &  62.30 &   \textbf{71.07} & 66.12 &   \textbf{66.50} \\
&MM+MIRO     &   \textbf{79.45} &   52.39 &   \textbf{78.15} &   \textbf{70.00} &   \textbf{63.45} &   \textbf{71.07} & 64.37 &   \textbf{66.30} \\
&MM+CL    &   \textbf{77.79} &   53.68 &   \textbf{79.25} &   \textbf{70.24} & 61.38 &   \textbf{71.47} & 66.84 &   \textbf{66.56} \\\midrule
\multirow{5}{*}{M2D} 
&MM+ERM        &   \textbf{80.25} &    \textbf{56.43} &   \textbf{78.81} &   \textbf{71.83} &  62.30 &   68.13 & 64.68 &    \textbf{65.04} \\
&MM+Mixup   &   \textbf{79.67} &   52.39 & 76.05 &   \textbf{69.37} & 60.00 &    \textbf{70.90} & 66.02 &   \textbf{65.51} \\
&MM+MMD   &   \textbf{80.03} &    \textbf{54.23} &   \textbf{79.03} &   \textbf{72.12} & 61.84 & 67.73 & 65.91 &   \textbf{65.01} \\
&MM+MIRO    &  75.72 &   \textbf{54.23} &   \textbf{82.91} &   70.95 & 60.92 & 66.27 & 65.40 & 64.20 \\
&MM+CL   &   \textbf{78.37} &    \textbf{54.14} &  76.82 &   \textbf{69.78} &  62.30 & 66.67 & 61.70 & 63.56 \\ \midrule[1pt]
\multicolumn{10}{c}{ViT-based Backbone} \\ \midrule[1pt]
\multirow{4}{*}{} 
&DeepAll   \cite{dong2023simmmdg}   & 91.56 & 61.12 & 83.97 & \uline{78.88}  & 57.47 & 64.93 & 61.19 & 61.20 \\
&SimMMDG    \cite{dong2023simmmdg}    & \uline{91.71} & 54.69 & \uline{84.22} & 76.87 & 55.86 & 66.53 & 61.19 & 61.19 \\
&CMRF  \cite{fan2024cross}    & 86.53 & \uline{66.36} & 83.71 & 78.87  & 57.70 & 66.00 & 62.42 & 62.04  \\ 
&MDJAT \cite{li2025towards} & 90.98 & 58.27 & 84.12 & 77.79 & 58.39 & \uline{67.86} & \uline{64.17} & \uline{63.47} \\ 
&MAD-DG \cite{ji2026alignment} & 80.09 & 58.04 & 79.79 & 72.64 & \uline{60.05} & 64.69 & 62.09 & 62.27 \\
\midrule
\multirow{5}{*}{D2M}
 & MM+ERM          &  90.05 & 51.75 &   \textbf{85.65} &  75.82 &    \textbf{61.15} &   \textbf{71.60} &    \textbf{66.43} &    \textbf{66.39} \\
 & MM+Mixup   & 87.17 &  60.20 &  80.7 &  76.02 &   \textbf{65.06} &   \textbf{72.93} &    \textbf{69.20} &   \textbf{69.06} \\
 & MM+MMD      &  85.65 & 59.10 &   \textbf{85.21} & 76.65 &   \textbf{63.22} &   \textbf{71.33} &    \textbf{66.22} &    \textbf{66.92} \\
 & MM+MIRO    &  87.53 & 54.69 & 80.35 & 74.19 &   \textbf{63.68} &   \textbf{72.67} &    \textbf{67.15} &   \textbf{67.83} \\
 & MM+CL      &  86.59 & 54.23 & 82.34 & 74.39 &   \textbf{64.60} &   \textbf{73.60} &    \textbf{68.99} &   \textbf{69.06} \\
\midrule
\multirow{5}{*}{M2D}
 & MM+ERM        &  90.92 &  62.13 &   \textbf{86.53} &   \textbf{79.86} &   \textbf{62.30} &   \textbf{70.27} &  62.94 &   \textbf{65.17} \\
 & MM+Mixup    &  91.28 & 57.54 &   \textbf{85.98} &  78.27 &   \textbf{62.30} &   \textbf{71.47} &  63.86 &   \textbf{65.88} \\
 & MM+MMD      &  90.99 &  63.14 &   \textbf{86.64} &   \textbf{80.26} &   \textbf{63.45} &   \textbf{68.67} &   \textbf{64.89} &   \textbf{65.67} \\
 & MM+MIRO     &  91.35 & 63.51 &   \textbf{85.54} &   \textbf{80.13} &   \textbf{61.84} &   \textbf{71.07} & 62.63 &   \textbf{65.18} \\
 & MM+CL      &  90.77 &  62.04 &   \textbf{85.87} &   \textbf{79.56} &   \textbf{62.99} &   \textbf{68.80} & 62.42 &   \textbf{64.74} \\
\bottomrule[1pt]
\end{tabular}
\end{table*}

\noindent \textbf{Results across modality combinations.}
Table~\ref{tab_hac_epic_all} provides results with video, audio, and optical flow on HAC and EPIC-Kitchens under both CNN- and ViT- based setting. Across these settings, the comparison reveals clear advantages of our MMDG variants, while showing consistent dataset-dependent preferences between D2M and M2D frameworks.

With CNN backbones, our variants clearly surpass the reproduced baselines on both datasets. On HAC, the best mean improves from 66.78 (MDJAT) to 72.12 (M2D+MMD), while on EPIC-Kitchens, the best mean improves from 64.80 (SimMMDG) to 66.98 (D2M+Mixup). These gains highlight the value of a systematic framework for integrating DG with MML, rather than relying on ad-hoc model design. 
For framework comparison, M2D is generally stronger on HAC with higher best mean and more consistent improvements, whereas D2M performs better on EPIC-Kitchens. This suggests the preferred framework can be task-dependent, but both provide clear benefits when paired with the unified MML setting and DG strategies.

Replacing CNNs with ViTs, the reproduced baselines show a dataset-dependent shift where performance rises markly on HAC but drops on EPIC-Kitchens. 
This dataset-dependent shift suggests that simply upgrading the backbone does not reliably improve generalization for methods that rely primarily on MML with limited or no DG.
In contrast, our MML+DG variants under both D2M and M2D consistently outperforms the best reproduced baselines with the same ViT backbones, showing that the gains are not merely architectural but stem from improved domain robustness. The effect is most pronounced on EPIC-Kitchens, where our methods reverse the baseline degradation, while on HAC we further extend the already-strong ViT baselines, demonstrating additive benefits under stable cross-modal structure.
Finally, the ViT setting reinforces our framework guideline: M2D is preferred on HAC, whereas D2M is more reliable on EPIC-Kitchens by reducing dependence on cross-modal correlations before fusion. Further discussion on framework selection is provided in Section~\ref{sec_discussion}.

\subsection{MMDG Face Anti-Spoofing} \label{sec_fas}
\textbf{Datasets.}
We evaluate MMDG-Bench in face anti-spoofing (FAS) on three datasets: CASIA-CeFA (C)~\cite{liu2021casia}, CASIA-SURF (S)~\cite{zhang2020casia}, and WMCA (W)~\cite{george2019biometric}.
All provide three modalities of RGB, Depth, and Infrared (IR) and include diverse presentation attack instruments (PAIs).
Each dataset is treated as a separate domain. We adopt leave-one-domain-out evaluation (e.g., W, S$\to$C; W,C$\to$S; C,S$\to$W), training on two and testing on the held-out one.
Following ~\cite{lin2024suppress}, performance is evaluated using Half Total Error Rate (HTER) and Area Under the ROC Curve (AUC), where lower HTER and higher AUC indicate better generalization. \\
\textbf{Baselines.}
We compare the ten MMDG variants under both frameworks against ViT-Base-CA~\cite{liu2021casia} and mmdg \cite{lin2024suppress}, and report reproduced results under a unified setup that matches our methods in backbone architecture and optimizer. \\ 
\textbf{Implementation.} 
Following prior works \cite{lin2024suppress,yu2023flexible}, ViT-Base pretrained on ImageNet is adopted as the backbone encoder. RGB, depth, and IR images are resized to 224×224×3, with single-channel Depth and IR inputs replicated across three channels to match the ViT format. Each input is divided into 14×14 patches (16×16 patch size) plus one class token, resulting in 197 tokens with a hidden dimension of 768. The final-layer class token is used as the modality-specific feature representation.
Since all modalities yield 768-dimensional features, no projection layers are required in the D2M and M2D frameworks.
Other components, including the bottleneck fusion module and DG classifiers, follow the configurations in Section~\ref{sec_sc}. The model is trained for 30 epochs using SAM strategy~\cite{foret2020sharpness} with AdamW ~\cite{loshchilov2017decoupled} as the base optimizer and a perturbation radius of $\rho{=}0.05$. The learning rate is set to $1{\times}10^{-4}$ with weight decay of $1{\times}10^{-4}$ and a batch size of 32. Cosine annealing scheduling decays the learning rate from its initial value to a minimum of $1{\times}10^{-6}$. Each experiment typically requires 6–9 hours on a single NVIDIA H100 (80 GB) GPU. Detailed dataset descriptions and additional hyper-parameters are presented in Supplementary Material.

Table~\ref{tab_fas_rgb_depth_ir} provides MMDG results with RGB, Depth and IR. Both D2M and M2D consistently outperform the reproduced baselines, yielding lower mean HTER and higher AUC. The best variants reduce mean HTER from 21.92/20.42 (ViT-Base-CA/mmdg) to 14.40 (M2D+MMD), while improving mean AUC to 90.05. Comparing the two frameworks, M2D achieves the superior performance, indicating that M2D framework captures invariant cross-modal relationships enhances feature discrimination and improves cross-domain generalization.

\begin{table*}[!htbp]
\centering
\setlength{\tabcolsep}{0.7pt}
\small
\caption{
MMDG results with RGB, Depth and IR for FAS.
\uline{Underline}: The best result among the re-implemented models for a fair comparison.  
\textbf{Bold}: The results surpassing the best-reproduced result in bold.
Results for two modality cases (RGB + Depth, RGB+IR, and Depth+IR) are provided in the Supplementary Material. 
}
\label{tab_fas_rgb_depth_ir}
\begin{tabular}{@{}l c rr rr rr rr@{}}
\toprule
\multirow{2}{*}{} & \multirow{2}{*}{Model} &   
\multicolumn{2}{c}{W,S$\rightarrow$C} & \multicolumn{2}{c}{W,C$\rightarrow$S} & \multicolumn{2}{c}{C,S$\rightarrow$W} & \multicolumn{2}{c}{Mean} \\
\cmidrule(lr){3-4} \cmidrule(lr){5-6} \cmidrule(lr){7-8} \cmidrule(lr){9-10}
   &  & HTER ↓ & AUC ↑ & HTER ↓ & AUC ↑ & HTER  ↓ & AUC ↑ & HTER  ↓ & AUC ↑ \\
\midrule
 & ViT-Base-CA   \cite{yu2023flexible}     & \uline{10.67} & \uline{96.01} & 13.01 & 94.37 & 42.09 & 62.66 & 21.92 & \uline{84.35} \\
 &mmdg    \cite{lin2024suppress}   & 11.48 & 95.18 & \uline{8.20} & \uline{97.30} & \uline{41.59} & \uline{59.01} & \uline{20.42} & 83.83 \\ \midrule
\multirow{5}{*}{D2M} & MM+ERM   &    \textbf{2.67} &   \textbf{98.91} &   \textbf{7.95} &   \textbf{97.51} &  45.48 &  57.47 &   \textbf{18.70} &   \textbf{84.63} \\
 &MM+Mixup  &     \textbf{4.08} &   \textbf{98.49} & 16.07 & 90.97 &   \textbf{36.45} &   \textbf{70.52} &   \textbf{18.87} &   \textbf{86.66}  \\
 &MM+MMD   &   \textbf{1.67} &   \textbf{99.73} & 10.22 & 93.29 & 55.47 & 45.84 & 22.45 & 79.62 \\
 &MM+MIRO  &    \textbf{3.83} &   \textbf{98.62} & 15.15 & 92.09 &   \textbf{35.40} &   \textbf{70.48} &   \textbf{18.13} &   \textbf{87.06}\\
 &MM+CL   &   \textbf{1.33} &   \textbf{99.74} & 11.29 & 94.64 &  42.37 &   \textbf{60.40} &   \textbf{18.33} &   \textbf{84.93} \\ \midrule
\multirow{5}{*}{M2D} 
&MM+ERM       &   \textbf{1.67} &   \textbf{99.84} & 14.24 & 92.95 &   \textbf{41.49} &   \textbf{65.38} &   \textbf{19.13} &   \textbf{86.06} \\
&MM+Mixup     &   \textbf{1.08} &   \textbf{99.91} & 11.22 & 95.17 &   \textbf{40.19} &	  \textbf{62.04}   &   \textbf{17.50} &   \textbf{85.71} \\
&MM+MMD     &   \textbf{3.67} &   \textbf{98.02} &   \textbf{6.89} & 96.94 &   \textbf{32.63} &   \textbf{75.20} &   \textbf{14.40} &   \textbf{90.05} \\
&MM+MIRO      &   \textbf{1.25} &   \textbf{99.92} & 12.88 & 93.08 &   \textbf{41.23} &   \textbf{64.00} &   \textbf{18.45} &   \textbf{85.67} \\
&MM+CL   &   \textbf{1.33} &   \textbf{99.92} & 16.72 & 91.20 &   \textbf{40.96} &   \textbf{65.52} &   \textbf{19.67} &   \textbf{85.55} \\ 
\bottomrule
\end{tabular}
\end{table*}

\subsection{Further Analyses} \label{sec_ana}
\begin{figure*}[!htbp]
\centering
\begin{subfigure}[t]{0.52\textwidth}
  \centering
  \includegraphics[width=\textwidth]{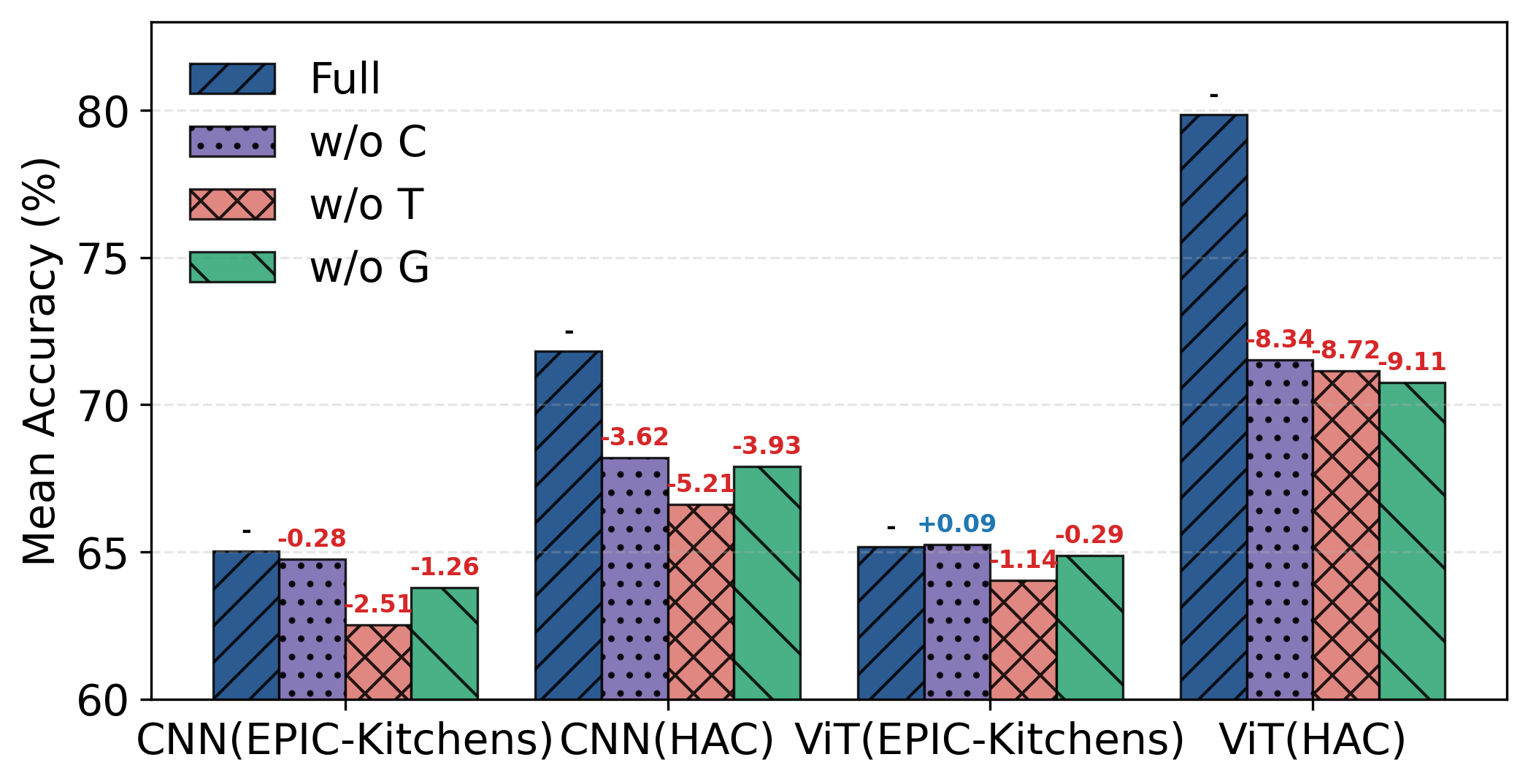}
  \caption{MM component ablation on EPIC-Kitchens and HAC with CNN and ViT-Base backbones under M2D framework.}
  \label{fig_aba_mm}
\end{subfigure} \hfill
\begin{subfigure}[t]{0.46\textwidth}
  \centering
  \includegraphics[width=\textwidth]{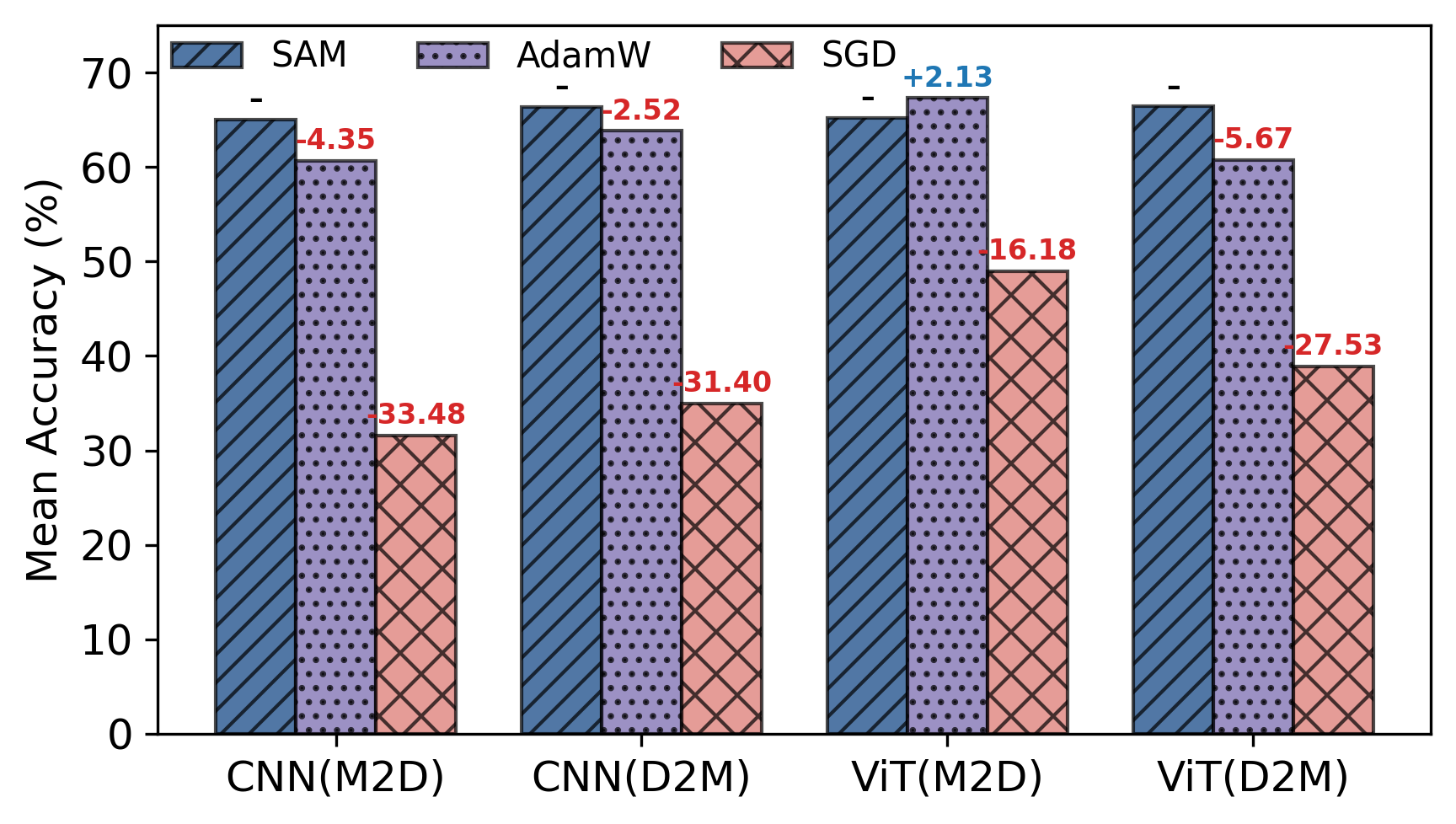}
  \caption{Optimizer comparison on EPIC-Kitchens under the M2D and D2M frameworks with CNN and ViT-Base backbones.}
  \label{fig_aba_optim}
\end{subfigure}
\caption{MM component ablation and optimizer comparison. Bars report the mean accuracy (\%) averaged over the three sub-tasks, and per-subtask results are provided in the Supplementary Material.}
\label{fig_aba_combined}
\end{figure*}

\textbf{MM components ablation.}
This section studies the role of the MM components, i.e., Supervised Contrastive Learning (C), Modality Translation (T), and Gradient Blending (G), by removing each from the simplest variant (MM+ERM, treated as the full configuration), yielding three ablations: w/o C, w/o T, and w/o G. The fusion module (F) is kept fixed to produce multi-modal predictions in all variants.
Experiments are conducted under M2D on EPIC-Kitchens and HAC with both CNN and ViT-Base backbones, following Section~\ref{sec_sc}, with results shown in Fig.~\ref{fig_aba_mm}. 
Removing MM components usually reduce performance, indicating that C/T/G provide complementary benefits when modal-gap mitigation is performed before domain learning. The only exception is ViT on EPIC-Kitchens, where removing C yields a slight improvement. This is likely because the supervised contrastive objective may over-collapse same-label features across modalities and interfere with fusion or translation.
Besides, the ablation sensitivity is dataset-dependent. On EPIC-Kitchens, ablations cause only modest drops. In contrast, on HAC, removing any component leads to larger degradation for both backbones, especially using ViT as the backbone, suggesting that HAC exhibits stronger modality imbalance and requires explicit cross-modal consistency and adaptive fusion for robust generalization.

\noindent \textbf{Optimizer comparison.} 
This section studies the impact of optimizer choice in MMDG-Bench by comparing SAM \cite{foret2020sharpness}, AdamW \cite{loshchilov2017fixing}, and SGD \cite{gower2019sgd} on the same base variant (MM+ERM) under both D2M and M2D, evaluated on EPIC-Kitchens with CNN and ViT-Base backbones following Section~\ref{sec_sc}. As shown in Fig.~\ref{fig_aba_optim}, optimizer choice has a clear effect and interacts with both backbone and framework order.
With CNN, SAM is consistently best under both frameworks (65.04 in M2D and 66.33 in D2M), while AdamW is weaker and SGD collapses to very low accuracy, likely due to its sensitivity to learning-rate/momentum schedules and cross-modality gradient-scale imbalance under our default hyperparameters.
With ViT-Base, SAM remains stable and competitive, while AdamW is order-sensitive, peaking in ViT–M2D (67.30) but dropping in ViT–D2M (60.72), suggesting a less favorable D2M-induced landscape for transformer optimization.
Overall, SAM is the most reliable across settings, whereas AdamW can be competitive but is less robust to the framework choice, and SGD is consistently inferior. We therefore adopt SAM as the default optimizer in this study.

\noindent \textbf{Evaluation with missing modalities} 
Real-world FAS systems may face sensor failures that make certain modalities unavailable.
To evaluate robustness, we evaluate three missing-modality settings at test time: missing Depth (RGB+IR), missing IR (RGB+Depth), and missing Depth \& IR (RGB only). All models are trained with full modalities and tested with incomplete inputs using the same configurations as in Section~\ref{sec_fas}. Table~\ref{tab_fas_missing_overall} reports mean HTER and AUC across all tasks.
Both D2M and M2D consistently outperform the reproduced baselines under test-time missing-modality conditions, with M2D generally achieving the strongest overall performance. Although all methods suffer degradation relative to the fixed-modality setting in Table~\ref{tab_fas_rgb_depth_ir}, the drop is much more severe for mmdg \cite{lin2024suppress}, whereas M2D variants, such as MM+MMD, remain more robust.
Besides, the performance gap between the two frameworks widens under missing-modality conditions: the best M2D model (MM+MMD) achieves 26.03\% HTER, outperforming the best D2M model (35.72\%).
This indicates that M2D in FAS encourages each modality to develop semantically complete and complementary representations that remain effective even when others are absent.

\begin{table*}[!htbp]
\centering
\small
\setlength{\tabcolsep}{0.7pt}
\caption{
Average HTER and AUC across three tasks (W,S $\to$C; W,C$\to$S; C,S$\to$W) in FAS under test-time missing-modality scenarios. \uline{Underline}: The best result among the re-implemented models for a fair comparison.  
\textbf{Bold}: The results surpassing the best-reproduced result in bold.
Per-task results are provided in the Supplementary Material.}
\label{tab_fas_missing_overall}
\begin{tabular}{llcccccccc}
\hline
\multirow{2}{*}{} & \multirow{2}{*}{Model} &
\multicolumn{2}{c}{Missing Depth} &
\multicolumn{2}{c}{Missing IR} &
\multicolumn{2}{c}{Missing Depth \& IR} &
\multicolumn{2}{c}{Mean} \\ 
\cmidrule(lr){3-4} \cmidrule(lr){5-6}  \cmidrule(lr){7-8}   \cmidrule(lr){9-10} 
 & & HTER ↓ & AUC ↑ & HTER↓ & AUC↑ & HTER↓ & AUC↑ & HTER↓ & AUC↑ \\
\hline
  \multicolumn{2}{c}{ViT-Base-CA\cite{yu2023flexible}}  &  \uline{44.06} & \uline{57.34} & 21.97 & 82.35 & \uline{40.69} & \uline{60.75} & \uline{35.57} & \uline{66.81} \\  
& mmdg \cite{lin2024suppress} & 49.43 & 51.10 & \uline{19.38} & \uline{83.00} & 63.63 & 35.31 & 44.14 & 56.47 \\
\hline
\multirow{5}{*}{D2M} 
& MM+ERM & 48.70 & 51.73 &   \textbf{17.72} &   \textbf{86.34} & 42.92 &   56.12 & 36.45 & 64.73 \\
 & MM+Mixup & 45.13 & 55.22 &   \textbf{18.67} &   \textbf{84.54} & 43.37 &   56.40 & 35.72 & 65.39 \\
 & MM+MMD & 48.73 & 51.92 &   \textbf{18.07} &   \textbf{86.00} &   43.04 &   55.51 & 36.62 & 64.48 \\
 & MM+MIRO & 48.83 & 51.51 &   \textbf{17.89} &   \textbf{86.31} & 42.78 &  56.06 & 36.50 & 64.63 \\
 & MM+CL & 48.73 & 51.81 &   \textbf{17.63} &   \textbf{86.37} & 42.97 &  56.11 & 36.44 & 64.77 \\
\hline
\multirow{5}{*}{M2D} 
 & MM+ERM &   \textbf{41.05} &   \textbf{57.87} &   \textbf{19.16} &   \textbf{83.38} &   \textbf{37.07} &   \textbf{61.61} &   \textbf{32.43} &   \textbf{67.62} \\
 & MM+Mixup &   \textbf{41.25} & 57.05 &   \textbf{16.93} &   \textbf{86.65} &   \textbf{38.30} &  60.34 &   \textbf{32.17} &   \textbf{68.02} \\
 & MM+MMD &   \textbf{32.26} &   \textbf{66.13} &   \textbf{14.23} &   \textbf{89.56} &   \textbf{31.59} &   \textbf{66.81} &   \textbf{26.03} &   \textbf{74.17} \\
 & MM+MIRO &   \textbf{43.68} & 56.19 &   \textbf{18.36} &   \textbf{85.50} &   \textbf{35.32} &   \textbf{63.78} &   \textbf{32.46} &   \textbf{68.49} \\
 & MM+CL &   \textbf{42.60} & 56.93 &   \textbf{19.09} &   \textbf{83.52} &   \textbf{39.18} &  60.38 &   \textbf{33.62} &   \textbf{66.94} \\
\hline
\end{tabular}
\end{table*}


\noindent \textbf{Cross-domain stability of cross-modal relations.}
This section quantifies how cross-modal relationships change across domains, providing evidence for when M2D or D2M is preferable. For each domain $d$ and modality $m$, frozen backbones are used to extract features and compute class centroids $A_d^m\in\mathbb{R}^{C\times D_m}$.
To compare modalities with different feature dimensions (e.g., audio vs.\ video/flow), a canonical correlation analysis \cite{knapp1978canonical}-inspired correlation operator is defined and instantiated on the prototypes to form a cross-modal relation matrix $M^{(d)}$, where $M^{(d)}[i,j]$ measures the affinity between class $i$ in modality $m_1$ and class $j$ in modality $m_2$. The full derivations of $A_d^m\in\mathbb{R}^{C\times D_m}$ and $M^{(d)}$ are provided in the Supplementary Material. Cross-domain stability of cross-modal relations is then quantified by CMRS, defined as the mean pairwise Spearman rank correlation \cite{zar1972significance} between vectorized relation matrices across all domain pairs:
 \(\text{CMRS}=\frac{1}{\binom{|\mathcal{D}|}{2}}\sum_{i<j}
\text{Spearman}\!\left(\text{vec}(M^{(d_i)}),\,\text{vec}(M^{(d_j)})\right).\)
Higher CMRS indicates more stable cross-modal relations across domains. 
For binary classification tasks such as FAS, where the two-class label space
renders the full $C{\times}C$ relation matrix degenerate and uninformative, we
introduce a binary-compatible extension, $\text{CMRS}_{\text{bin}}$.
Given the relation matrix $M^{(d)}$ for domain $d$, we define a discriminative
affinity score $s^{(d)} = \frac{M^{(d)}_{11}+M^{(d)}_{22}}{2}
         -\frac{M^{(d)}_{12}+M^{(d)}_{21}}{2}$, 
which contrasts within-class against between-class cross-modal affinities, and
compute stability as $\text{CMRS}_{\text{bin}}
= 1 - \frac{\operatorname{std}_d\!\left[s^{(d)}\right]}
           {\operatorname{mean}_d\!\left[s^{(d)}\right]}$.
Higher $\text{CMRS}_{\text{bin}}$ indicates that the discriminative cross-modal
structure is more consistent across domains.

\begin{figure*}[!htbp]
  \centering
  \begin{minipage}[t]{0.53\textwidth}
    \centering
    \vspace{0pt}
    \captionof{table}{CMRS scores across domain and modality pairs.}
    \label{tab_detailed_cmrs}
    \small  
    \begin{tabular}{lcccc}
      \toprule
      \textbf{HAC} & H--C & H--A & A--C &  Mean $\pm$ Std  \\ \midrule
      A--F & 0.898 & 0.644 & 0.559 & $0.700 \pm 0.176$ \\
      V--A & 0.862 & 0.706 & 0.563 & $0.710 \pm 0.150$ \\
      V--F & 0.799 & 0.763 & 0.852 & $0.805 \pm 0.045$ \\ \midrule
      \textbf{EPIC} & 1--2 & 1--3 & 2--3 &  Mean $\pm$ Std  \\ \midrule
      A--F & 0.829 & 0.838 & 0.738 & $0.802 \pm 0.055$ \\
      V--A & 0.745 & 0.802 & 0.698 & $0.748 \pm 0.052$ \\
      V--F & 0.799 & 0.933 & 0.761 & $0.831 \pm 0.090$ \\ \bottomrule
    \end{tabular}
  \end{minipage}
  \hfill
  \begin{minipage}[t]{0.46\textwidth}
    \centering
    \vspace{0pt}
    \includegraphics[width=\textwidth]{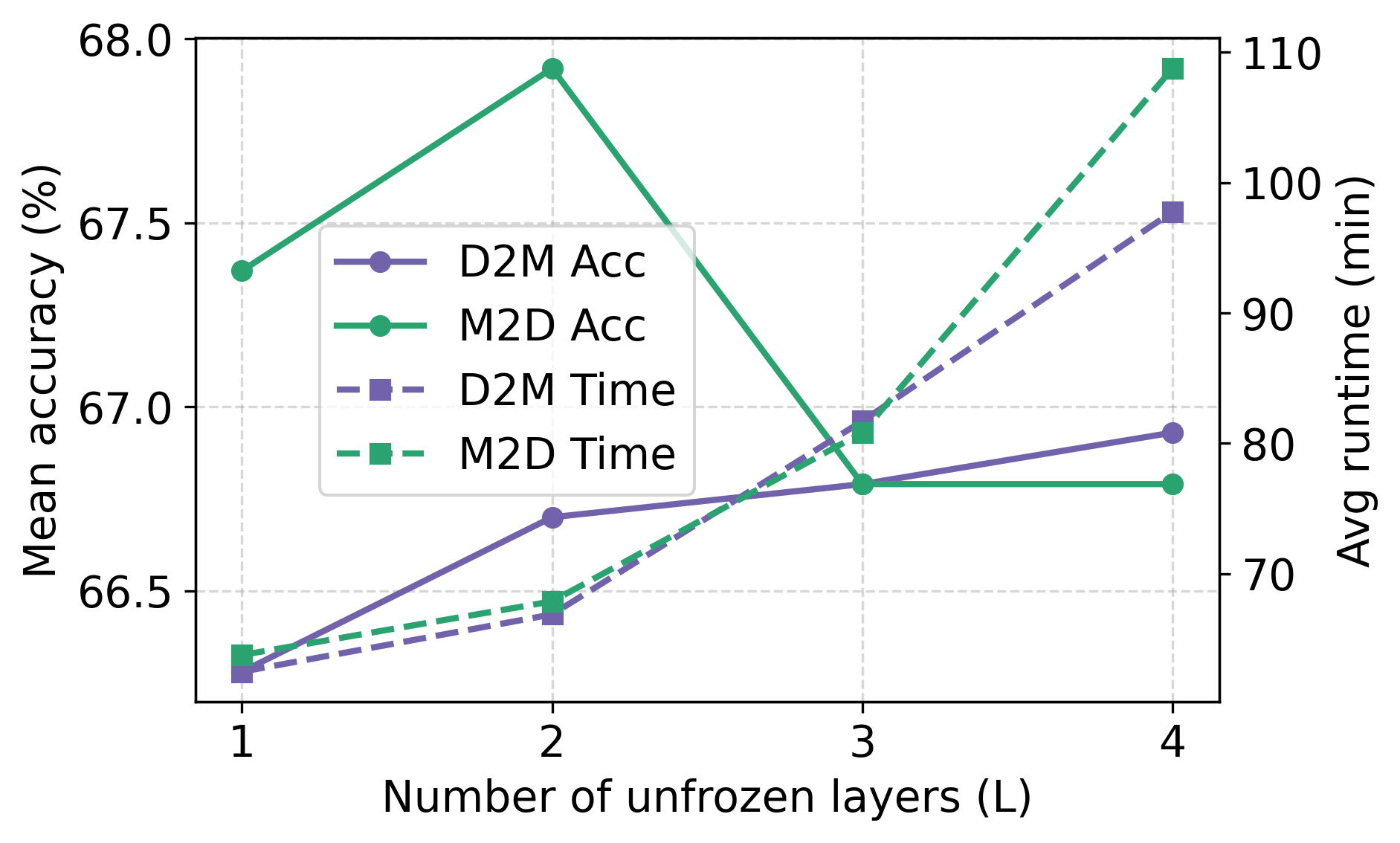}
    \caption{Effect of unfreezing layers ($L$) for MM+ERM on HAC under D2M/M2D.}
    \label{fig_ana_layer}
  \end{minipage}
\end{figure*}

Cross-modal stability is evaluated on HAC and EPIC-Kitchens using three
modalities: video (V), audio (A), and optical flow (F), with CMRS reported
in Table~\ref{tab_detailed_cmrs}. For FAS, where the binary label space
renders the standard CMRS degenerate, we instead report $\text{CMRS}_{\text{bin}}$
per modality pair. On HAC, the visual pair V–F is the most stable
($0.805\pm0.045$), whereas audio-related pairs are less stable and more uneven
across domains, with a notable drop for A–F from H–C (0.898) to A–C (0.559).
On EPIC-Kitchens, CMRS is consistently high and relatively uniform across
kitchen domains for all modality pairs, indicating that the cross-modal
structure is largely preserved across domains. EPIC-Kitchens (D1/D2/D3)
involves the same cooking activity across different kitchens, so domain shifts
are largely environmental and preserve cross-modal semantics, leading to high
and uniform CMRS. In contrast, HAC spans ontology-level domains
(human/cartoon/animal) with different audio characteristics, which can change
the audio–visual/flow mapping and yield lower, more uneven CMRS for
audio-involved pairs. For FAS, $\text{CMRS}_{\text{bin}}$ scores are
\{0.897, 0.694, 0.081\} for \{RGB--Depth, RGB--IR, Depth--IR\} pairs
(mean: 0.557), indicating low and uneven cross-modal stability overall.
Accordingly, high and uniform CMRS (or $\text{CMRS}_{\text{bin}}$) favours
D2M, whereas lower or uneven values favour M2D (Insight~\ding{173},
Section~\ref{sec_discussion}). This is consistent with M2D outperforming D2M
on FAS, further validating the generality of the framework-selection guideline
across both multi-class and binary settings.

\begin{figure*}[!htbp]
\centering
\includegraphics[width=\textwidth]{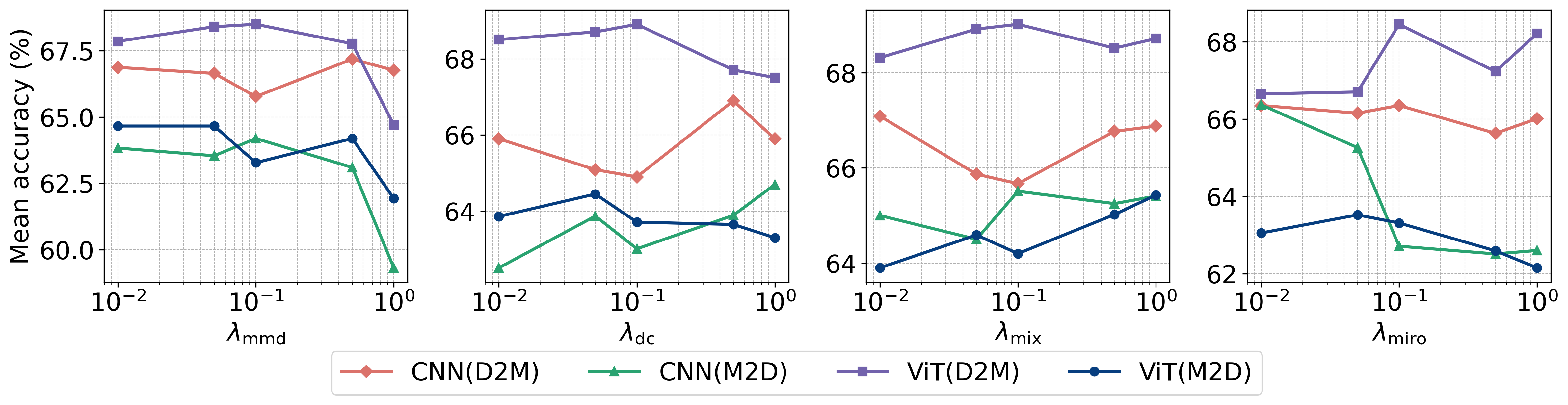}
\caption{Sensitivity of DG regularization weights $\lambda_{mmd}$, $\lambda_{dc}$ $\lambda_{mix}$ and $\lambda_{miro}$ for MM+MMD, MM+CL, MML+ Mixup, MML+MIRO on EPIC-Kitchens, sweep over $\{0.01, 0.05, 0.1, 0.5, 1.0\}$. Curves report mean accuracy averaged across the three sub-tasks under D2M/M2D with CNN and ViT-Base. Results for per-subtask results are provided in the Supplementary Material.
}
\label{fig_ana_para}
\end{figure*}

\noindent \textbf{Parameter analyses.} This section analyzes the sensitivity of the DG regularization weights $\lambda_{mmd}$, $\lambda_{dc}$, $\lambda_{mix}$, $\lambda_{miro}$ in MM+MMD, MM+CL, MM+Mixup, and MM+MIRO. We evaluate EPIC-Kitchens under both D2M and M2D with CNN and ViT-Base by sweeping each weight over ${0.01, 0.05, 0.1, 0.5, 1.0}$ while keeping all other settings the same as Section~\ref{sec_sc}. Fig.~\ref{fig_ana_para} reports the mean accuracy over the three EPIC-Kitchens sub-tasks, with per-subtask results in the Supplementary Material. The four weights show moderate sensitivity: performance changes smoothly and remains competitive across a broad mid-range, indicating that the conclusions are not dependent on precise tuning. 
Specifically, $\lambda_{\mathrm{mmd}}$ and $\lambda_{\mathrm{dc}}$ are stable over $0.01$--$0.5$, while overly large values (especially $1.0$) may cause over-regularization. $\lambda_{\mathrm{mix}}$ shows a clearer optimum around $0.1$, whereas $\lambda_{\mathrm{miro}}$ is more sensitive but still performs best in the mid-range ($0.1$--$0.5$), with no benefit at extreme values. Based on these trends, we set $\{\lambda_{\mathrm{mmd}}, \lambda_{\mathrm{dc}}, \lambda_{\mathrm{mix}}, \lambda_{\mathrm{miro}}\}=\{0.05,\,0.05,\,0.1,\,0.5\}$ on EPIC-Kitchens, which lies in empirically stable or near-optimal regions.

\noindent \textbf{Effect of backbone layer unfreezing.}
This section studies fine-tuning depth by varying the number of unfrozen CNN backbone layers ($L$=1–4). At $L=1$, only the top stage and classifier are trained. Deeper settings progressively unfreeze lower blocks.
Unfrozen backbone parameters are trained with a reduced learning rate ($0.5\times10^{-4}$), whereas classification heads/fusion modules use $1\times10^{-4}$. All other settings follow Section~\ref{sec_sc}.
Fig.~\ref{fig_ana_layer} reports the mean accuracy and runtime on HAC across four modality combinations using MM+ERM under both frameworks, with EPIC-Kitchens results in the Supplementary Material.
Overall, deeper unfreezing yields only marginal accuracy changes but noticeably increases training time. D2M varies only slightly from 66.28\% at $L=1$ to 66.93\% at 
$L=4$, while M2D reaches its best mean accuracy at $L=2$ (67.92\%) and then plateaus at 66.79\% for $L=3$ and $L=4$. In contrast, runtime increases substantially, from about 1h2m at $L=1$ to 1h37m–1h49m at $L=4$. These results suggest that $L=1$ or $L=2$ offers the best accuracy–efficiency trade-off, with $L=2$ being the most balanced choice overall.

\section{Discussion}
\label{sec_discussion}
Evaluation across action recognition and FAS tasks shows that our MMDG variants under D2M and M2D often surpass previous methods, highlighting the importance of unified benchmarking for advancing this field. Beyond performance, we derive several key insights, discussed below.

\textbf{Insight \ding{172}: Incorporating DG is key to boost generalization.}
Across action recognition (Table \ref{tab_hac_epic_all}) and FAS (Tables \ref{tab_fas_rgb_depth_ir}), our MML+DG variants under both D2M and M2D consistently outperform prior MMDG methods that rely mainly on MML with limited or no DG, indicating that MML alone is insufficient for generalization.
More importantly, upgrading CNNs to stronger ViTs (Table~\ref{tab_hac_epic_all}) yields dataset-dependent changes for non-DG baselines (gaining on HAC but degrading on EPIC-Kitchens), whereas our variants remain consistently beneficial and often amplify the gains from stronger pretraining. 
The same stabilizing effect is also observed in Table~\ref{tab_fas_missing_overall}, where variants with explicit DG remain more robust under missing modalities.
This contrast highlights DG as the stabilizer to make the resulting representations more reliable and robust.

\textbf{Insight \ding{173}: Choose the framework based on cross-modal relationship stability.}
D2M performs better on EPIC-Kitchens, while M2D is more reliable on HAC (Table \ref{tab_hac_epic_all}).
We attribute this difference to the quantified cross-domain stability of cross-modal relations (CMRS, Table \ref{tab_detailed_cmrs}).  Specifically, higher and more uniform CMRS indicates that the cross-modal mapping is largely stable across domains, while lower or uneven CMRS suggests that the mapping itself varies with domains. This criterion aligns naturally with the two framework choice. D2M aligns each modality across domains before fusion, and is therefore preferable when the cross-modal mapping is stable across domains (as in EPIC-Kitchens, where different kitchens share similar cross-modal semantics), since domain alignment mainly removes environmental nuisance factors and facilitates subsequent fusion. In contrast, M2D performs cross-modal alignment/fusion within each domain first, which is safer when the cross-modal mapping is domain-dependent (as in HAC with ontology-level shifts across human/cartoon/animal), because early modality-wise domain alignment in D2M can force mismatched correspondences. M2D instead leverages domain-specific cross-modal complementarity before enforcing domain invariance, leading to more reliable generalization.

\textbf{Insight \ding{174}: Stronger backbones raise the ceiling and sharpen the framework contrast.}
Replacing CNNs with ViTs (Table~\ref{tab_hac_epic_all}) in MMDG-Bench improves overall accuracy and widens the performance gap between D2M and M2D.
Stronger pretraining makes cross-modal cues easier to exploit, thereby amplifying the role of cross-domain correspondence stability. D2M becomes more advantageous when cross-modal correspondences are stable across domains, as in EPIC-Kitchens, whereas M2D is more reliable when such correspondences are domain-dependent, as in HAC.
As a result, framework choice is more critical with modern backbones.

\section{Conclusion}
\label{sec_conclusion}
This study introduce MMDG-Bench, a comprehensive benchmark comprising two complementary frameworks: D2M and M2D, with unified protocols and flexible integration of diverse MML and DG methods across tasks. Extensive experiments on action recognition and face anti-spoofing demonstrate that our ten MMDG variants often surpass prior state-of-the-art methods, highlighting the critical need for standardized benchmarking in advancing MMDG research.
Beyond performance gains, our analysis yields key insights for framework selection, DG integration, and practical deployment. This work establishes a principled foundation for MMDG through unified frameworks, rigorous benchmarking, and actionable guidelines.
 
\clearpage  


%
%
\bibliographystyle{splncs04}
\bibliography{main}
\end{document}